\documentclass[11pt,letterpaper]{article}
\usepackage{naaclhlt2016}
\usepackage{times}
\usepackage{latexsym}
\usepackage{graphicx}
\usepackage{color}
\usepackage{booktabs}
\usepackage[dvipsnames]{xcolor}
\usepackage{subcaption}
\usepackage{multirow}
\usepackage{verbatim}
\usepackage{fancyhdr}
\usepackage{amsmath}
\naaclfinalcopy
\naaclfinalcopy 
\def\naaclpaperid{639} 
\usepackage{url}
\usepackage{xspace}
\newcommand{\JHUAff}[0]{\ensuremath{2}\xspace}
\newcommand{\URAff}[0]{\ensuremath{3}\xspace}
\newcommand{\CMUAff}[0]{\ensuremath{1}\xspace}
\newcommand{\VTAff}[0]{\ensuremath{4}\xspace}
\newcommand{\MSRAff}[0]{\ensuremath{6}\xspace}
\newcommand{\FBAff}[0]{\ensuremath{5}\xspace}

\definecolor{light-gray}{gray}{0.95}

\usepackage[font=small,labelfont=bf]{caption}
\renewcommand{\belowcaptionskip}{-2.5mm}
\renewcommand{\abovecaptionskip}{1mm}
\begin{document}

\title{Visual Storytelling}

\author{
Ting-Hao (Kenneth) Huang$^\CMUAff$\thanks{\ \ T.H. and F.F. contributed equally to this work.}, 
Francis Ferraro$^{\JHUAff}$\footnotemark[1],
Nasrin Mostafazadeh$^{\URAff}$, Ishan Misra$^\CMUAff$, Aishwarya Agrawal$^\VTAff$, \\ {\bf Jacob Devlin$^\MSRAff$, Ross Girshick$^\FBAff$, Xiaodong He$^\MSRAff$, Pushmeet Kohli$^\MSRAff$, Dhruv Batra$^\VTAff$, C.~Lawrence Zitnick$^\FBAff$,} \\ {\bf Devi Parikh$^\VTAff$, Lucy Vanderwende$^\MSRAff$, Michel Galley$^\MSRAff$, Margaret Mitchell$^\MSRAff$} \vspace{.25cm} \\ 
{\bf Microsoft Research} \\
{\small $\CMUAff$ Carnegie Mellon University,} 
{\small $\JHUAff$ Johns Hopkins University,} 
{\small $\URAff$ University of Rochester,} \\
{\small $\VTAff$ Virginia Tech,}
{\small $\FBAff$ Facebook AI Research} \\
{\small $\MSRAff$} Corresponding authors: \{jdevlin,lucyv,mgalley,memitc\}@microsoft.com \\
}


\maketitle

\begin{abstract}
We introduce the first dataset for {\bf sequential vision-to-language}, and explore how this data may be used for the task of {\it visual storytelling}.  The first release of this dataset, SIND\footnote{Sequential Images Narrative Dataset.  This and future releases are made available on {\tt sind.ai}.}~v.1, includes  81,743 unique photos in 20,211 sequences, aligned to both descriptive (caption) and story language.  We establish several strong baselines for the storytelling task, and motivate an automatic metric to benchmark progress. Modelling concrete description as well as figurative and social language, as provided in this dataset and the storytelling task, has the potential to move artificial intelligence from basic understandings of typical visual scenes towards more and more human-like understanding of grounded event structure and subjective expression.
\end{abstract}

\section{Introduction}
\begin{figure}[t]
\centering
\includegraphics[width=0.48\textwidth]{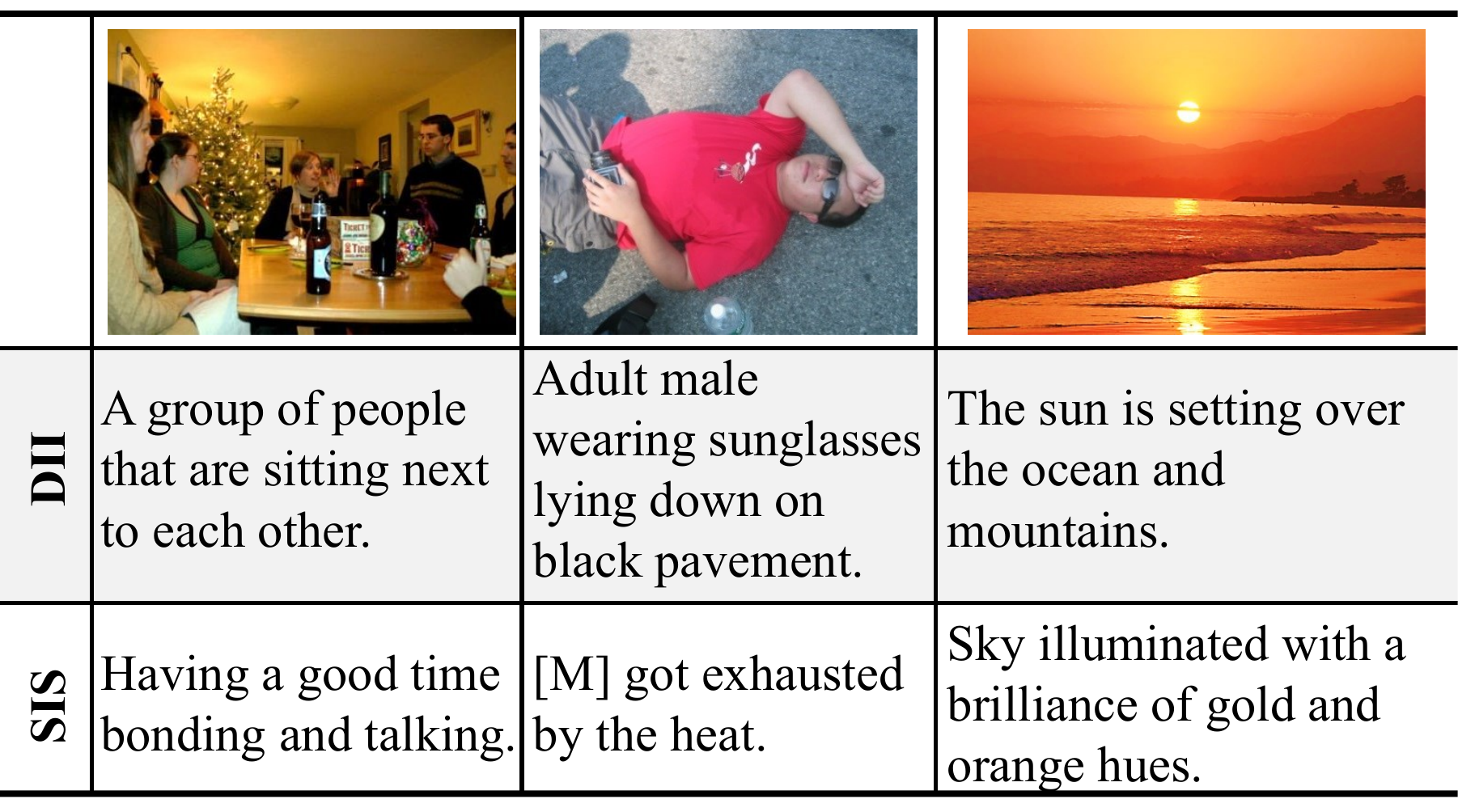}
\caption{Example language difference between descriptions for images in isolation (DII) vs.~stories for images in sequence (SIS).}
\label{fig:alternations}
\end{figure}

Beyond understanding simple objects and concrete scenes lies interpreting causal structure; making sense of visual input to tie disparate moments together as they give rise to a cohesive narrative of events through time.  This requires moving from reasoning about single images -- static moments, devoid of context -- to sequences of images that depict events as they occur and change.  On the vision side, progressing from single images to images in context allows us to begin to create an artificial intelligence (AI) that can reason about a visual moment given what it has already seen.  
On the language side, progressing from literal description to narrative helps to learn more evaluative, conversational, and abstract language.  This is the difference between, for example, ``sitting next to each other'' versus ``having a good time'', or ``sun is setting'' versus ``sky illuminated with a brilliance...'' (see Figure \ref{fig:alternations}).  The first descriptions capture image content that is literal and concrete; the second requires further inference about what a {\it good time} may look like, or what is special and worth sharing about a particular sunset.




We introduce the first dataset of sequential images with corresponding descriptions,  
 which captures some of these subtle but important differences, and advance the task of visual storytelling.  We 
 release the data in three tiers of language for the same images:  \textbf{(1) Descriptions of images-in-isolation (DII)}; {\bf (2) Descriptions of images-in-sequence (DIS)}; and {\bf (3) Stories for images-in-sequence (SIS)}.  This tiered approach reveals the effect of temporal context and the effect of narrative language.  As all the tiers are aligned to the same images, the dataset facilitates directly modeling the relationship between literal and more abstract visual concepts, including the relationship between visual imagery and typical event patterns.  We additionally propose an automatic evaluation metric which is best correlated with human judgments, and establish several strong baselines for the visual storytelling task.


\section{Motivation and Related Work}

Work in vision to language has exploded, with researchers examining image captioning~\cite{lin2014microsoft,Karpathy_2015_CVPR,VinyalsCvpr2015,Xu2015show,chen2015deja,young2014image,elliott2013visualdependency}, question answering~\cite{VQA,NIPS2015_5640,NIPS2015_5641,malinowski2014nips}, visual phrases~\cite{sadeghi2011recognition}, video understanding~\cite{ramanathan2013video}, and visual concepts~\cite{krishnavisualgenome,fangCVPR15}.

Such work focuses on direct, literal description of image content.  While this is an encouraging first step in connecting vision and language, it is far from the capabilities needed by intelligent agents for naturalistic interactions. 
There is a significant difference, yet unexplored, between remarking that a visual scene shows ``sitting in a room'' -- typical of most image captioning work -- and that the same visual scene shows ``bonding''.  The latter description is grounded in the visual signal, yet it brings to bear information about social relations and emotions that can be additionally inferred in context (Figure~\ref{fig:alternations}).  
Visually-grounded stories facilitate more evaluative and figurative language than has previously been seen in vision-to-language research:  If a system can recognize that colleagues look {\it bored}, it can remark and act on this information directly.  

Storytelling itself is one of the oldest known human activities \cite{wiessner2014embers}, 
providing a way to educate, preserve culture, instill morals, and share advice; focusing AI research towards this task therefore has the potential to bring about more human-like intelligence and understanding.

\section{Dataset Construction}



\begin{table}[t]
\centering
\small
\resizebox{\columnwidth}{!}{
\begin{tabular}{@{}lll@{}}
beach (684) & breaking up (350) & easter (259) \\
amusement park (525) & carnival (331) &church (243) \\
building a house (415) & visit (321) & graduation ceremony (236) \\
party (411) & market (311) & office (226) \\
birthday (399) & outdoor activity (267) & father's day (221) \\
\end{tabular}}
\caption{The number of albums in our tiered dataset for the 15 most frequent kinds of stories.}
\label{tab:query-terms}
\end{table}

\paragraph{Extracting Photos}
We begin by generating a list of ``storyable'' event types. 
We leverage the idea that ``storyable'' events tend to involve some form of possession, e.g., ``John\textit{'s} birthday party,'' or ``Shabnam\textit{'s} visit.'' %
Using the Flickr data release~\cite{yfcc100m}, we aggregate 5-grams of photo titles and descriptions, using Stanford CoreNLP~\cite{manning-corenlp-2014} to extract possessive dependency patterns. %
We keep the heads of possessive phrases if they can be classified as an \textsc{event} in WordNet3.0, relying on manual winnowing to target our collection efforts.\footnote{
We simultaneously supplemented this data-driven effort by a small hand-constructed gazetteer. %
}
These terms are then used to collect albums using the Flickr API.\footnote{https://www.flickr.com/services/api/}
We only include albums with 10 to 50 photos where all album photos are taken within a 48-hour span and CC-licensed.  See Table~\ref{tab:query-terms} for the query terms with the most albums returned.

\begin{figure}[t]
\centering
\includegraphics[width=0.48\textwidth]{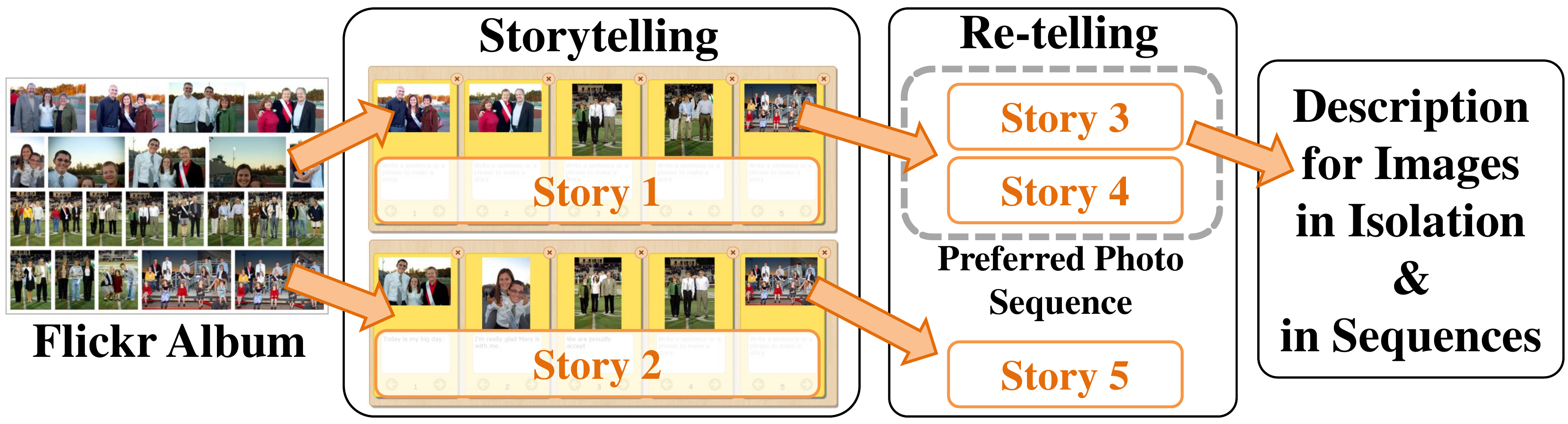}
\caption{Dataset crowdsourcing workflow.}
\label{fig:workflow}
\end{figure}

The photos returned from this stage are then presented to crowd workers using Amazon's Mechanical Turk to collect the corresponding stories and descriptions.  The crowdsourcing workflow of developing the complete dataset is shown in Figure~\ref{fig:workflow}.

%



\paragraph{Crowdsourcing Stories In Sequence}

\begin{figure}[t]
\centering
\includegraphics[width=0.48\textwidth]{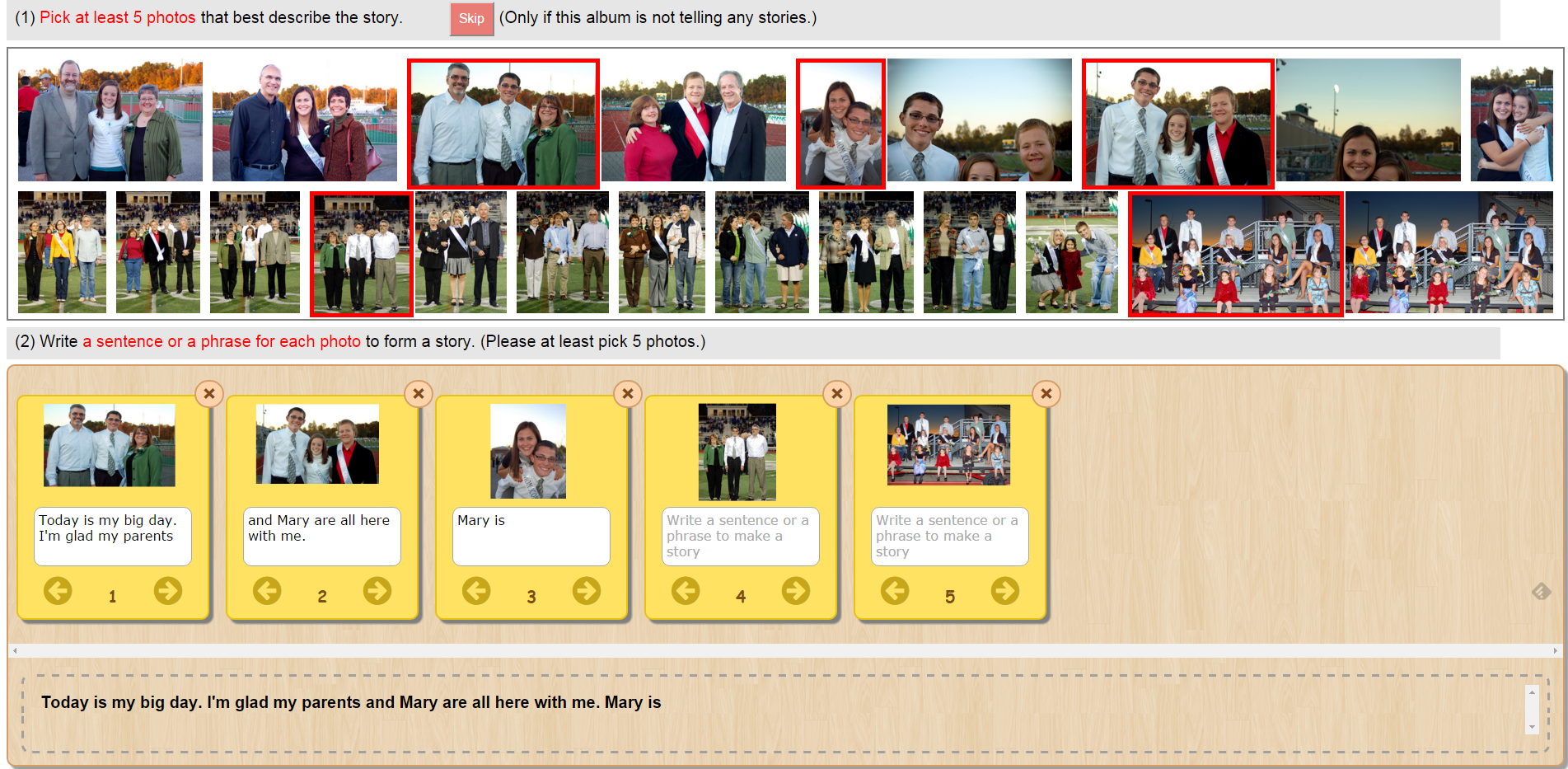}
\caption{Interface for the \textit{Storytelling} task, which contains: 1) the photo album, and 2) the storyboard.}
\label{fig:ui}
\end{figure}

We develop a 2-stage crowdsourcing workflow to collect naturalistic stories with text aligned to images.  The first stage is \textit{storytelling}, where the crowd worker selects a subset of photos from a given album to form a photo sequence and writes a story about it (see Figure~\ref{fig:ui}).
The second stage is \textit{re-telling}, in which the worker writes a story based on one photo sequence generated by workers in the first stage.

In both stages, all album photos are displayed in the order of the time that the photos were taken, with a ``storyboard'' underneath.
In \textit{storytelling}, by clicking a photo in the album, a ``story card'' of the photo appears on the storyboard.
The worker is instructed to pick at least five photos, arrange the order of selected photos, and then write a sentence or a phrase on each card to form a story; this appears as a full story underneath the text aligned to each image. Additionally, this interface captures the alignments between text and photos. Workers may skip an album if it does not seem storyable (e.g., a collection of coins).  Albums skipped by two workers are discarded.
The interface of \textit{re-telling} is similar, but it displays the two photo sequences already created in the first stage, which the worker chooses from to write the story.  For each album, 2 workers perform \textit{storytelling} (at \$0.3/HIT), and 3 workers perform \textit{re-telling} (at \$0.25/HIT), 
yielding a total of 1,907 workers.  All HITs use quality controls to ensure varied text at least 15 words long.





\paragraph{Crowdsourcing Descriptions of Images In Isolation \& Images In Sequence}

\begin{figure}[t]
\centering
\includegraphics[width=.48\textwidth]{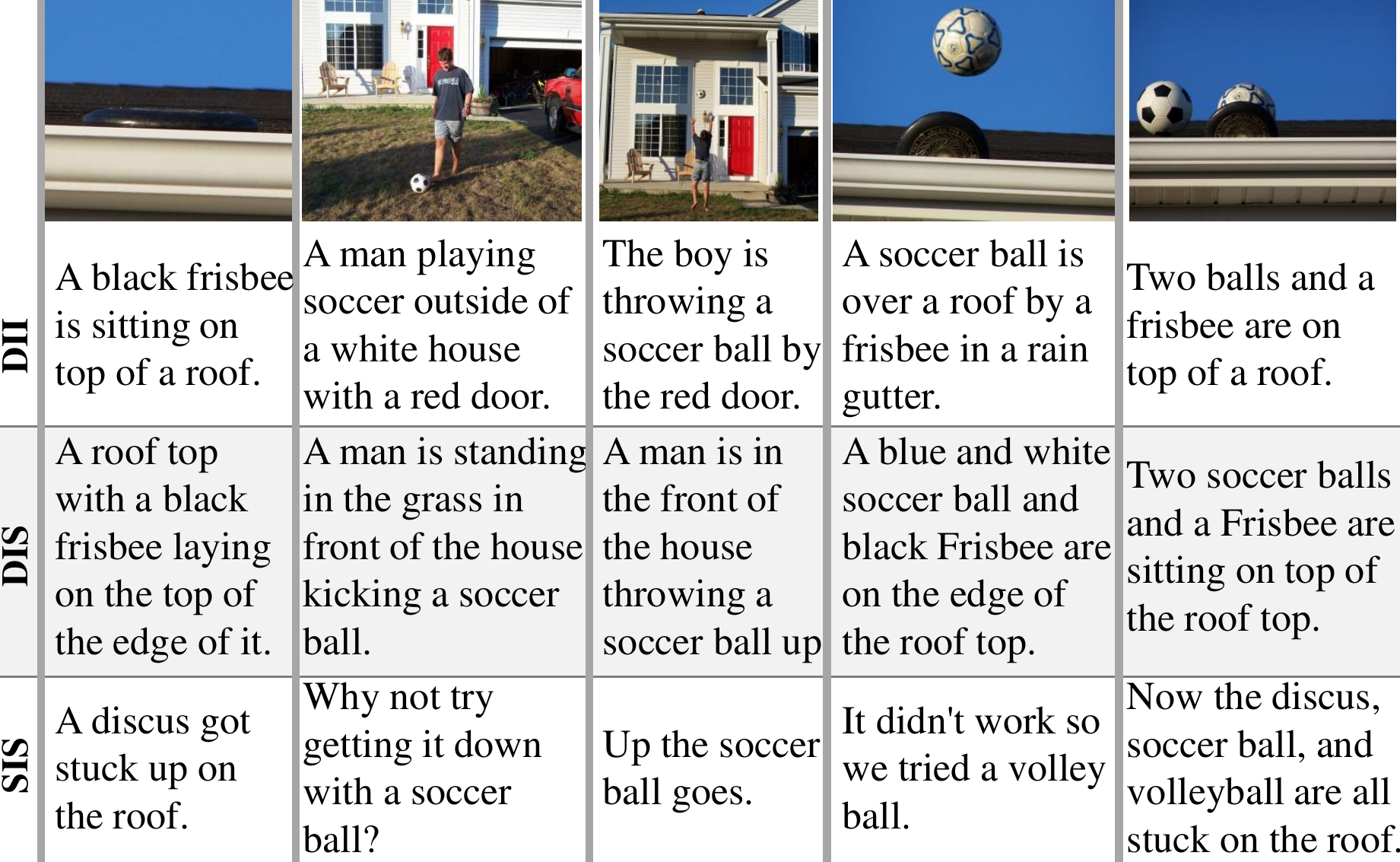}
\caption{Example descriptions of images in isolation (DII); descriptions of images in sequence (DIS); and stories of images in sequence (SIS).}
\label{fig:story-example}
\end{figure}

We also use crowdsourcing to collect  descriptions of images-in-isolation (DII) and descriptions of images-in-sequence (DIS), for the photo sequences with stories from a majority of workers in the first task (as Figure~\ref{fig:workflow}). 
In both DII and DIS tasks, workers are asked to follow the instructions for image captioning proposed in MS COCO~\cite{lin2014microsoft} such as \textit{describe all the important parts}. 
In DII, we use the MS COCO image captioning interface.\footnote{https://github.com/tylin/coco-ui} 
In DIS, we use the storyboard and story cards of our \textit{storytelling} interface to display a photo sequence, with MS COCO instructions adapted for sequences. We recruit 3 workers for DII (at \$0.05/HIT) and 3 workers for DIS (at \$0.07/HIT). 


\paragraph{Data Post-processing}
We tokenize all storylets and descriptions with the CoreNLP tokenizer,  %
and replace all people names with generic \textsc{male}/\textsc{female} tokens,\footnote{
We use those names occurring at least 10,000 times. \url{https://ssa.gov/oact/babynames/names.zip}
} %
and all identified named entities with their entity type (e.g., {\tt location}).  The data is released as {\it training}, {\it validation}, and {\it test} following an 80\%/10\%/10\% split on the stories-in-sequence albums.  Example language from each tier is shown in Figure \ref{fig:story-example}.

\section{Data Analysis}

\begin{table}[t]
  \includegraphics[width=0.48\textwidth]{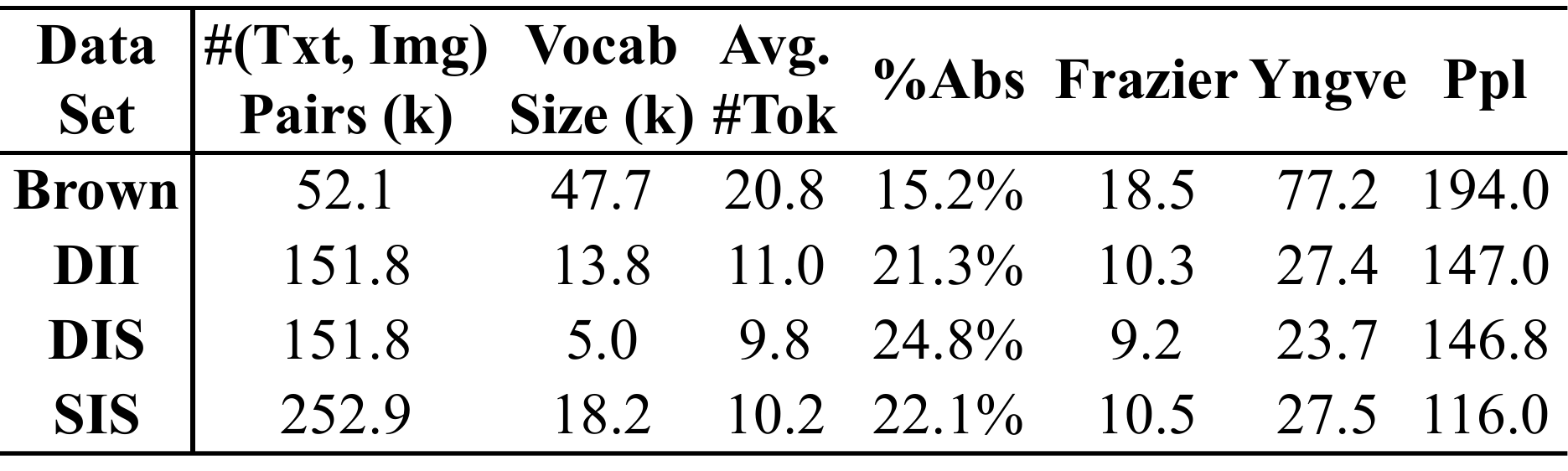}
  \caption[]{A summary of our dataset, following the proposed analyses of~\newcite{FerraroEtAl15}, including the Frazier and Yngve measures of syntactic complexity. The balanced Brown corpus~\cite{BrownCorpus}, provided for comparison, contains only text.  Perplexity (Ppl) is calculated against a 5-gram language model learned on a generic 30B English words dataset scraped from the web.}
\label{tab:stats}
\end{table}

Our dataset includes 10,117 Flickr albums with 210,819 unique photos.
Each album on average has 20.8 photos ($\sigma$ = 9.0).
The average time span of each album is 7.9 hours ($\sigma$ = 11.4).
Further details of each tier of the dataset are shown in Table~\ref{tab:stats}.\footnote{We exclude words seen only once.}

\begin{table}
\includegraphics[width=.48\textwidth]{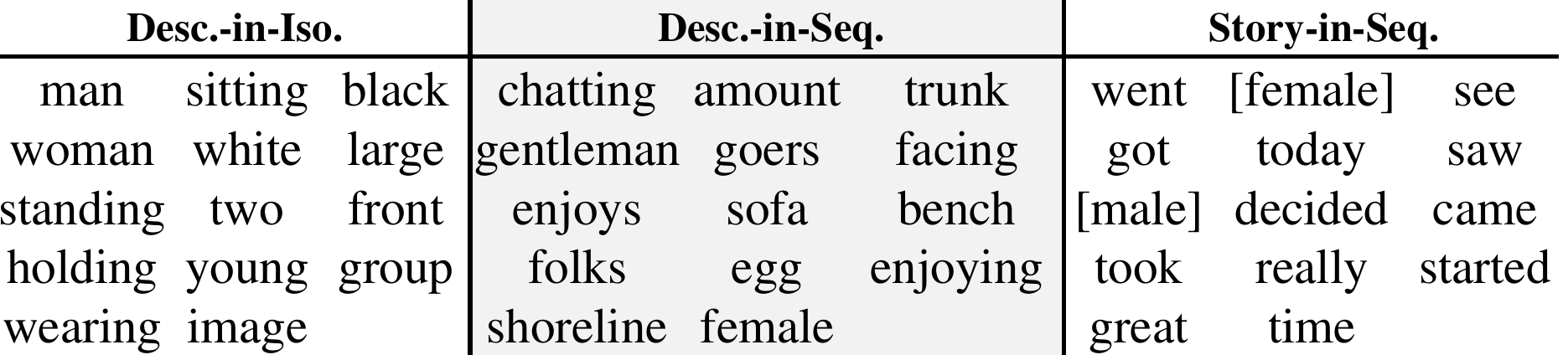}
\caption{Top words ranked by normalized PMI.}
\label{tab:vocab}
\end{table}

We use normalized pointwise mutual information to identify the words most closely associated with each tier (Table~\ref{tab:vocab}). Top words for descriptions-in-isolation reflect an impoverished disambiguating context: References to people often lack social specificity, as people are referred to as simply ``man'' or ``woman''. Single images often do not convey much information about underlying events or actions, which leads to the abundant use of posture verbs (``standing'', ``sitting'', etc.). As we turn to descriptions-in-sequence, these relatively uninformative words are much less represented. Finally, top story-in-sequence words include more storytelling elements, such as names ({\it $[$male$]$}), temporal references ({\it today}) and words that are more dynamic and abstract ({\it went}, {\it decided}).

\section{Automatic Evaluation Metric} 

Given the nature of the complex storytelling task, the best and most reliable evaluation for assessing the quality of generated stories is human judgment. 
However, automatic evaluation metrics are useful to quickly benchmark progress.  To better understand which metric could serve as a proxy for human evaluation, we compute pairwise correlation coefficients between automatic metrics and human judgments on 3,000 stories sampled from the SIS training set.

For the human judgements, we again use crowdsourcing on MTurk, asking five judges per story to rate how strongly they agreed with the statement ``If these were my photos, I would like using a story like this to share my experience with my friends''.\footnote{Scale presented ranged from ``Strongly disagree'' to ``Strongly agree'', which we convert to a scale of 1 to 5.}  We take the average of the five judgments as the final score for the story. 
 For the automatic metrics, we use METEOR,\footnote{We use METEOR version 1.5 with {\tt hter} weights.}  smoothed-BLEU \cite{smoothed_bleu}, and Skip-Thoughts~\cite{SkipThought}   
 to compute similarity between each story for a given sequence. Skip-thoughts provide a Sentence2Vec embedding which models the semantic space of novels.

\begin{table}
\centering
\small
\begin{tabular}{l|lll}
\hline
                 & METEOR            & BLEU              & Skip-Thoughts     \\ \hline
$r$  & 0.22 (2.8e-28) & 0.08 (1.0e-06) & 0.18 (5.0e-27) \\
$\rho$  & 0.20 (3.0e-31)  & 0.08 (8.9e-06) & 0.16 (6.4e-22) \\
$\tau$ & 0.14 (1.0e-33) & 0.06 (8.7e-08) & 0.11 (7.7e-24) \\ \hline
\end{tabular}
\caption{Correlations of automatic scores against human judgements, with p-values in parentheses.}
\label{tab:metric_correlations}
\end{table}

 As Table \ref{tab:metric_correlations} shows, METEOR correlates best with human judgment according to all the correlation coefficients. This signals that a metric such as METEOR which incorporates paraphrasing 
 correlates best with human judgement on this task. A more detailed study of automatic evaluation of stories is an area of interest for a future work.  



\section{Baseline Experiments}




\begin{table*}
\centering
\small
    \begin{tabular}{ccccc}
    \includegraphics[width=2.75cm]{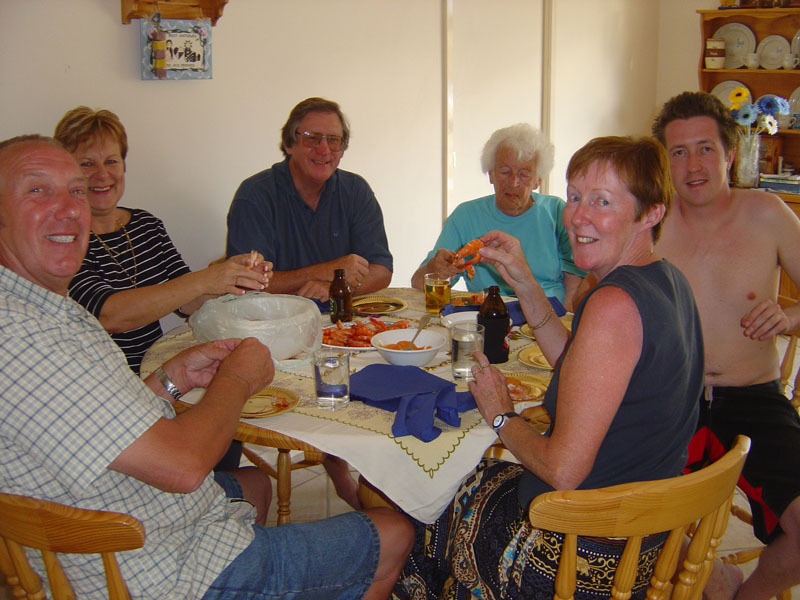} & 
    \includegraphics[width=2.75cm]{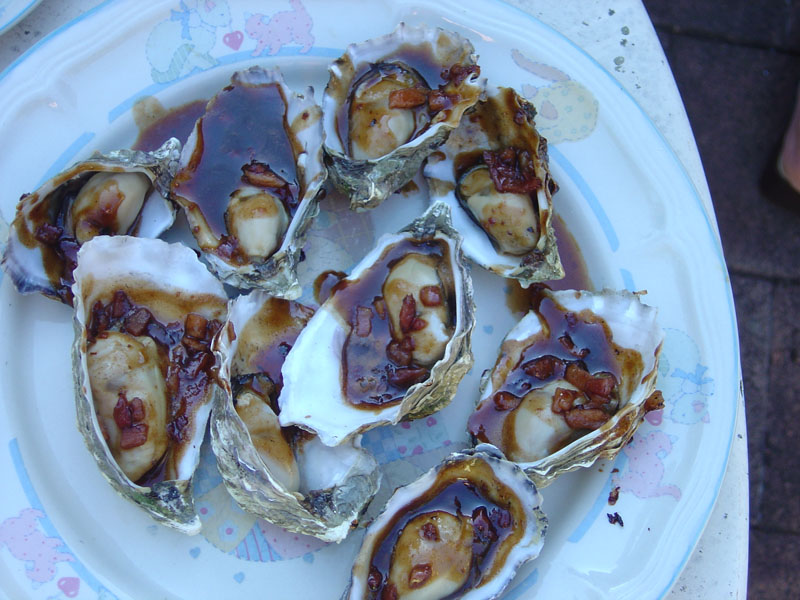} & 
    \includegraphics[width=2.75cm]{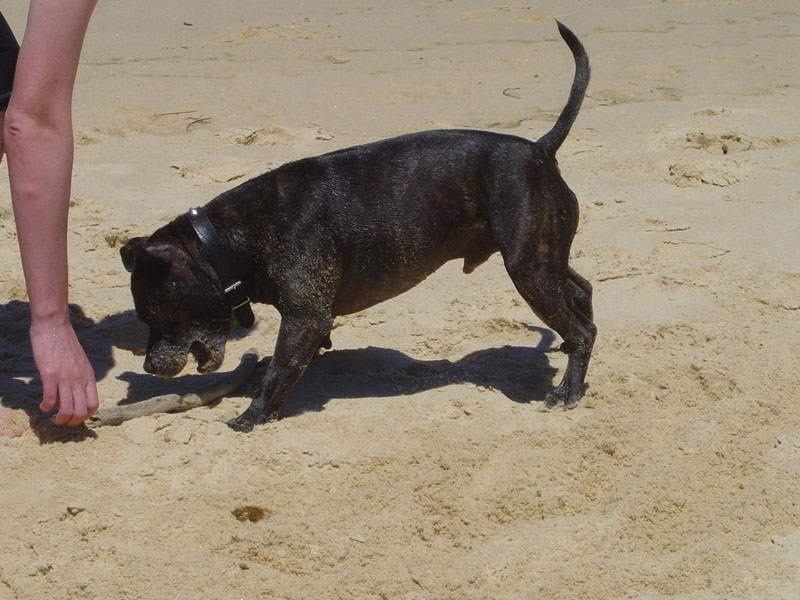} &
    \includegraphics[width=2.75cm]{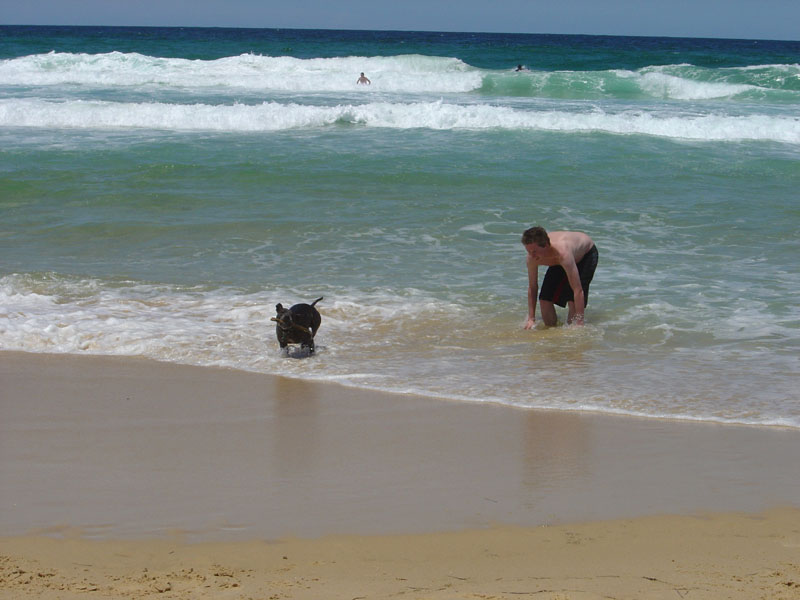} & 
    \includegraphics[width=2.75cm]{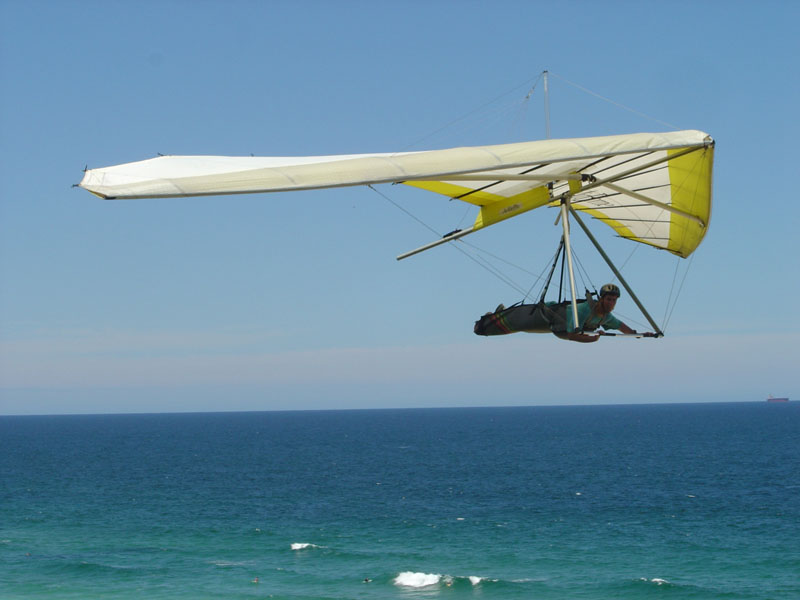} \\
    \end{tabular} \\
%
\small
    \begin{tabular}{rp{0.8\textwidth}}
    {\it +Viterbi} & {\small This is a picture of a family.  This is a picture of a cake.  This is a picture of a dog.  This is a picture of a beach. This is a picture of a beach.} \vspace{0.25em}\\
    {\it +Greedy} & {\small The family gathered together for a meal. The food was delicious. The dog was excited to be there. The dog was enjoying the water. The dog was happy to be in the water.} \vspace{0.25em}\\
    {\it -Dups} & {\small The family gathered together for a meal. The food was delicious. The dog was excited to be there. The kids were playing in the water. The boat was a little too much to drink.} \vspace{0.25em}\\
    {\it +Grounded} & {\small The family got together for a cookout. They had a lot of delicious food. The dog was happy to be there. They had a great time on the beach. They even had a swim in the water.} \\
    \end{tabular}
    \caption{Example stories generated by baselines.}
    \label{tab:example_stories}
\end{table*}

\begin{table}
\centering
\begin{tabular}{l|lll}
{\bf Beam=10} & {\bf Greedy} & {\bf -Dups} &  {\bf +Grounded} \\\hline 
23.55 & 19.10 & 19.21 & -- \\\hline
\end{tabular}
\caption{Captions generated per-image with METEOR scores.}
\label{tab:rnn_caption_baselines}
\end{table}

\begin{table}
\centering
\begin{tabular}{l|lll}

{\bf Beam=10} & {\bf Greedy} & {\bf -Dups} &  {\bf +Grounded} \\\hline 
23.13  & 27.76 & 30.11  & 31.42 \\
\end{tabular}
\caption{Stories baselines with METEOR scores.}
\label{tab:rnn_story_baselines}
\end{table}

We report baseline experiments on the storytelling task in Table \ref{tab:rnn_story_baselines}, training on the SIS tier and testing on half the SIS validation set (valtest).  Example output from each system is presented in Table \ref{tab:example_stories}.  To highlight some differences between story and caption generation, we also train on the DII tier in isolation, and produce captions per-image, rather than in sequence.  These results are shown in Table \ref{tab:rnn_story_baselines}.

To train the story generation model, we use a sequence-to-sequence recurrent neural net (RNN) approach, which naturally extends the single-image captioning technique of \newcite{DevlinEtAl15} and \newcite{Vinyals2015} to multiple images. Here, we encode an image {\it sequence} by running an RNN over the {\tt fc7} vectors of each image, in reverse order. This is used as the initial hidden state to the story decoder model, which learns to produce the story one word at a time using softmax loss over the training data vocabulary. We use Gated Recurrent Units (GRUs) \cite{Cho2014Arxiv} for both the image encoder and story decoder.

In the baseline system, we generate the story using a simple beam search (size=10), which has been successful in image captioning previously  \cite{DevlinEtAl15}. However, for story generation, the results of this model subjectively appear to be very poor -- the system produces generic, repetitive, high-level descriptions (e.g., ``This is a picture of a dog'').  This is a predictable result given the label bias problem inherent in maximum likelihood training; recent work has looked at ways to address this issue directly \cite{LiEtAl2016}.

To establish a stronger baseline, we explore several decode-time heuristics to improve the quality of the generated story. The first heuristic 
is to lower the decoder beam size substantially.  We find that using a beam size of 1 (greedy search) significantly increases the story quality, resulting in a 4.6 gain in METEOR score. However, the same effect is not seen for caption generation, with the greedy caption model obtaining worse quality than the beam search model.  This highlights a key difference in generating stories versus generating captions.


Although the stories produced using a greedy search result in significant gains, they include many repeated words and phrases, e.g., ``The kids had a great time. And the kids had a great time.'' We introduce a very simple heuristic to avoid this, where the same content word cannot be produced more than once within a given story. This improves METEOR by another 2.3 points.

An advantage of comparing captioning to storytelling side-by-side is that the captioning output may be used to help inform the storytelling output.  To this end, we include an additional baseline 
where ``visually grounded'' words may only be produced if they are licensed by the caption model. We define the set of visually grounded words to be those which occurred at higher frequency in the caption training than the story training: 
\vspace{-.5em}
\begin{equation}
\centering
\displaystyle\frac{P(w|T_{caption})}{P(w|T_{story})} > 1.0
\end{equation}
\vspace{-1em}

We train a separate model using the caption annotations, and produce an n-best list of captions for each image in the valtest set.  Words seen in at least 10 sentences in the 100-best list are marked as `licensed' by the caption model. Greedy decoding without duplication proceeds with the additional constraint that if a word is visually grounded, it can only be generated by the story model if it is licensed by the caption model for the same photo set. This results in a further 1.3 METEOR improvement.

It is interesting to note what a strong effect relatively simple heuristics have on the generated stories.  We do not intend to suggest that these heuristics are the {\it right} way to approach story generation. Instead, the main purpose is to provide clear baselines that demonstrate that story generation has fundamentally different challenges from caption generation; and the space is wide open to explore for training and decoding methods to generate fluent stories.

\section{Conclusion and Future Work}

We have introduced the first dataset for {\bf sequential vision-to-language}, which incrementally moves from images-in-isolation to stories-in-sequence. We argue that modelling the more figurative and social language captured in this dataset is essential for evolving AI towards more human-like understanding.  We have established several strong baselines for the task of visual storytelling, and have motivated METEOR as an automatic metric to evaluate progress on this task moving forward.


\bibliography{bib}

\begin{thebibliography}{}

\bibitem[\protect\citename{Antol \bgroup et al.\egroup }2015]{VQA}
Stanislaw Antol, Aishwarya Agrawal, Jiasen Lu, Margaret Mitchell, Dhruv Batra,
  C.~Lawrence Zitnick, and Devi Parikh.
\newblock 2015.
\newblock Vqa: Visual question answering.
\newblock In {\em International Conference on Computer Vision (ICCV)}.

\bibitem[\protect\citename{Chen \bgroup et al.\egroup }2015]{chen2015deja}
Jianfu Chen, Polina Kuznetsova, David Warren, and Yejin Choi.
\newblock 2015.
\newblock D\'{e}j\`{a} image-captions: A corpus of expressive descriptions in
  repetition.
\newblock In {\em Proceedings of the 2015 Conference of the North American
  Chapter of the Association for Computational Linguistics: Human Language
  Technologies}, pages 504--514, Denver, Colorado, May--June. Association for
  Computational Linguistics.

\bibitem[\protect\citename{Cho \bgroup et al.\egroup }2014]{Cho2014Arxiv}
Kyunghyun Cho, Bart van Merrienboer, Caglar Gulcehre, Fethi Bougares, Holger
  Schwenk, and Yoshua Bengio.
\newblock 2014.
\newblock Learning phrase representations using {RNN} encoder-decoder for
  statistical machine translation.
\newblock {\em CoRR}.

\bibitem[\protect\citename{Devlin \bgroup et al.\egroup }2015]{DevlinEtAl15}
Jacob Devlin, Hao Cheng, Hao Fang, Saurabh Gupta, Li~Deng, Xiaodong He,
  Geoffrey Zweig, and Margaret Mitchell.
\newblock 2015.
\newblock Language models for image captioning: The quirks and what works.
\newblock In {\em Proceedings of the 53rd Annual Meeting of the Association for
  Computational Linguistics and the 7th International Joint Conference on
  Natural Language Processing (Volume 2: Short Papers)}, pages 100--105,
  Beijing, China, July. Association for Computational Linguistics.

\bibitem[\protect\citename{Elliott and
  Keller}2013]{elliott2013visualdependency}
Desmond Elliott and Frank Keller.
\newblock 2013.
\newblock Image description using visual dependency representations.
\newblock In {\em Proceedings of the 2013 Conference on Empirical Methods in
  Natural Language Processing}, pages 1292--1302, Seattle, Washington, USA,
  October. Association for Computational Linguistics.

\bibitem[\protect\citename{Fang \bgroup et al.\egroup }2015]{fangCVPR15}
Hao Fang, Saurabh Gupta, Forrest~N. Iandola, Rupesh Srivastava, Li~Deng, Piotr
  Doll{\'{a}}r, Jianfeng Gao, Xiaodong He, Margaret Mitchell, John~C. Platt,
  C.~Lawrence Zitnick, and Geoffrey Zweig.
\newblock 2015.
\newblock From captions to visual concepts and back.
\newblock In {\em Computer Vision and Pattern Recognition (CVPR)}.

\bibitem[\protect\citename{Ferraro \bgroup et al.\egroup }2015]{FerraroEtAl15}
Francis Ferraro, Nasrin Mostafazadeh, Ting-Hao~K. Huang, Lucy Vanderwende,
  Jacob Devlin, Michel Galley, and Margaret Mitchell.
\newblock 2015.
\newblock A survey of current datasets for vision and language research.
\newblock In {\em Proceedings of the 2015 Conference on Empirical Methods in
  Natural Language Processing}, pages 207--213, Lisbon, Portugal, September.
  Association for Computational Linguistics.

\bibitem[\protect\citename{Gao \bgroup et al.\egroup }2015]{NIPS2015_5641}
Haoyuan Gao, Junhua Mao, Jie Zhou, Zhiheng Huang, Lei Wang, and Wei Xu.
\newblock 2015.
\newblock Are you talking to a machine? dataset and methods for multilingual
  image question.
\newblock In C.~Cortes, N.D. Lawrence, D.D. Lee, M.~Sugiyama, R.~Garnett, and
  R.~Garnett, editors, {\em Advances in Neural Information Processing Systems
  28}, pages 2287--2295. Curran Associates, Inc.

\bibitem[\protect\citename{Karpathy and Fei-Fei}2015]{Karpathy_2015_CVPR}
Andrej Karpathy and Li~Fei-Fei.
\newblock 2015.
\newblock Deep visual-semantic alignments for generating image descriptions.
\newblock In {\em The IEEE Conference on Computer Vision and Pattern
  Recognition (CVPR)}, June.

\bibitem[\protect\citename{Kiros \bgroup et al.\egroup }2015]{SkipThought}
Ryan Kiros, Yukun Zhu, Ruslan~R Salakhutdinov, Richard Zemel, Raquel Urtasun,
  Antonio Torralba, and Sanja Fidler.
\newblock 2015.
\newblock Skip-thought vectors.
\newblock In C.~Cortes, N.D. Lawrence, D.D. Lee, M.~Sugiyama, R.~Garnett, and
  R.~Garnett, editors, {\em Advances in Neural Information Processing Systems
  28}, pages 3276--3284. Curran Associates, Inc.

\bibitem[\protect\citename{Krishna \bgroup et al.\egroup
  }2016]{krishnavisualgenome}
Ranjay Krishna, Yuke Zhu, Oliver Groth, Justin Johnson, Kenji Hata, Joshua
  Kravitz, Stephanie Chen, Yannis Kalanditis, Li-Jia Li, David~A Shamma,
  Michael Bernstein, and Li~Fei-Fei.
\newblock 2016.
\newblock Visual genome: Connecting language and vision using crowdsourced
  dense image annotations.

\bibitem[\protect\citename{Li \bgroup et al.\egroup }2016]{LiEtAl2016}
Jiwei Li, Michel Galley, Chris Brockett, Jianfeng Gao, and Bill Dolan.
\newblock 2016.
\newblock A diversity-promoting objective function for neural conversation
  models.
\newblock {\em NAACL HLT 2016}.

\bibitem[\protect\citename{Lin and Och}2004]{smoothed_bleu}
Chin-Yew Lin and Franz~Josef Och.
\newblock 2004.
\newblock Automatic evaluation of machine translation quality using longest
  common subsequence and skip-bigram statistics.
\newblock In {\em Proceedings of the 42Nd Annual Meeting on Association for
  Computational Linguistics}, ACL '04, Stroudsburg, PA, USA. Association for
  Computational Linguistics.

\bibitem[\protect\citename{Lin \bgroup et al.\egroup }2014]{lin2014microsoft}
Tsung-Yi Lin, Michael Maire, Serge Belongie, James Hays, Pietro Perona, Deva
  Ramanan, Piotr Doll{\'a}r, and C~Lawrence Zitnick.
\newblock 2014.
\newblock Microsoft coco: Common objects in context.
\newblock In {\em Computer Vision--ECCV 2014}, pages 740--755. Springer.

\bibitem[\protect\citename{Malinowski and Fritz}2014]{malinowski2014nips}
Mateusz Malinowski and Mario Fritz.
\newblock 2014.
\newblock A multi-world approach to question answering about real-world scenes
  based on uncertain input.
\newblock In Z.~Ghahramani, M.~Welling, C.~Cortes, N.D. Lawrence, and K.Q.
  Weinberger, editors, {\em Advances in Neural Information Processing Systems
  27}, pages 1682--1690. Curran Associates, Inc.

\bibitem[\protect\citename{Manning \bgroup et al.\egroup
  }2014]{manning-corenlp-2014}
Christopher~D. Manning, Mihai Surdeanu, John Bauer, Jenny Finkel, Steven~J.
  Bethard, and David McClosky.
\newblock 2014.
\newblock The {Stanford} {CoreNLP} natural language processing toolkit.
\newblock In {\em Proceedings of 52nd Annual Meeting of the Association for
  Computational Linguistics: System Demonstrations}, pages 55--60.

\bibitem[\protect\citename{Marcus \bgroup et al.\egroup }1999]{BrownCorpus}
Mitchell Marcus, Beatrice Santorini, Mary~Ann Marcinkiewicz, and Ann Taylor.
\newblock 1999.
\newblock Brown corpus, treebank-3.

\bibitem[\protect\citename{Ramanathan \bgroup et al.\egroup
  }2013]{ramanathan2013video}
Vignesh Ramanathan, Percy Liang, and Li~Fei-Fei.
\newblock 2013.
\newblock Video event understanding using natural language descriptions.
\newblock In {\em Computer Vision (ICCV), 2013 IEEE International Conference
  on}, pages 905--912. IEEE.

\bibitem[\protect\citename{Ren \bgroup et al.\egroup }2015]{NIPS2015_5640}
Mengye Ren, Ryan Kiros, and Richard Zemel.
\newblock 2015.
\newblock Exploring models and data for image question answering.
\newblock In C.~Cortes, N.D. Lawrence, D.D. Lee, M.~Sugiyama, R.~Garnett, and
  R.~Garnett, editors, {\em Advances in Neural Information Processing Systems
  28}, pages 2935--2943. Curran Associates, Inc.

\bibitem[\protect\citename{Sadeghi and Farhadi}2011]{sadeghi2011recognition}
Mohammad~Amin Sadeghi and Ali Farhadi.
\newblock 2011.
\newblock Recognition using visual phrases.
\newblock In {\em Computer Vision and Pattern Recognition (CVPR), 2011 IEEE
  Conference on}, pages 1745--1752. IEEE.

\bibitem[\protect\citename{Thomee \bgroup et al.\egroup }2015]{yfcc100m}
Bart Thomee, David~A Shamma, Gerald Friedland, Benjamin Elizalde, Karl Ni,
  Douglas Poland, Damian Borth, and Li-Jia Li.
\newblock 2015.
\newblock The new data and new challenges in multimedia research.
\newblock {\em arXiv preprint arXiv:1503.01817}.

\bibitem[\protect\citename{Vinyals \bgroup et al.\egroup }2014]{Vinyals2015}
Oriol Vinyals, Alexander Toshev, Samy Bengio, and Dumitru Erhan.
\newblock 2014.
\newblock Show and tell: a neural image caption generator.
\newblock In {\em CVPR}.

\bibitem[\protect\citename{Vinyals \bgroup et al.\egroup
  }2015]{VinyalsCvpr2015}
Oriol Vinyals, Alexander Toshev, Samy Bengio, and Dumitru Erhan.
\newblock 2015.
\newblock Show and tell: A neural image caption generator.
\newblock In {\em Computer Vision and Pattern Recognition}.

\bibitem[\protect\citename{Wiessner}2014]{wiessner2014embers}
Polly~W Wiessner.
\newblock 2014.
\newblock Embers of society: Firelight talk among the ju/’hoansi bushmen.
\newblock {\em Proceedings of the National Academy of Sciences},
  111(39):14027--14035.

\bibitem[\protect\citename{Xu \bgroup et al.\egroup }2015]{Xu2015show}
Kelvin Xu, Jimmy Ba, Ryan Kiros, Kyunghyun Cho, Aaron Courville, Ruslan
  Salakhutdinov, Richard Zemel, and Yoshua Bengio.
\newblock 2015.
\newblock Show, attend and tell: Neural image caption generation with visual
  attention.
\newblock {\em arXiv preprint arXiv:1502.03044}.

\bibitem[\protect\citename{Young \bgroup et al.\egroup }2014]{young2014image}
Peter Young, Alice Lai, Micah Hodosh, and Julia Hockenmaier.
\newblock 2014.
\newblock From image descriptions to visual denotations: New similarity metrics
  for semantic inference over event descriptions.
\newblock {\em Transactions of the Association for Computational Linguistics},
  2:67--78.

\end{thebibliography}
\bibliographystyle{naacl2016}

\end{document}